\documentclass{llncs}

\usepackage{graphicx}
\usepackage{amsmath}
\usepackage{hyperref}
\usepackage[numbers]{natbib}
\usepackage{tcolorbox}
\usepackage{xcolor}
\usepackage{subcaption}
\usepackage{caption}
\usepackage{appendix}

\newtcolorbox{mycolorbox}[1][]{
    colframe=black!50,
    colback=green!5,
    coltitle=black,
    boxrule=0.75mm,
    width=\textwidth,
    arc=4mm,
    auto outer arc,
    #1
}

\newtcolorbox{yellowcolorbox}[1][]{
    colframe=black!50,
    colback=yellow!5,
    coltitle=black,
    boxrule=0.75mm,
    width=\textwidth,
    arc=4mm,
    auto outer arc,
    #1
    }

\usepackage{caption}
\usepackage{booktabs}  % Per tabelle più belle

\hypersetup{
    colorlinks=true,
    linkcolor=blue,
    citecolor=blue,
    filecolor=blue,
    urlcolor=blue
}

\begin{document}

\title{\Large More is More: \\ Addition Bias in Large Language Models}

\author{\large \textbf{Luca Santagata,\inst{1}
Cristiano De Nobili\inst{2, 3}}}

\institute{Department of Information Engineering and Computer Science, \\ \centering University of Trento, Italy. \\ 
\email{luca.santagata@studenti.unitn.it}\\ 
\vspace{1.5em}
\and
MHPC (SISSA/ICTP), Trieste, Italy.\\
\and
Pi School, Rome, Italy.\\
\email{cdenobi@sissa.it}}

\maketitle
\begin{abstract}
    
In this paper, we investigate the presence of additive bias in Large Language Models (LLMs), drawing a parallel to the cognitive bias observed in humans where individuals tend to favor additive over subtractive changes \citep{adams2021people}. Using a series of controlled experiments, we tested various LLMs, including GPT-3.5 Turbo, Claude 3.5 Sonnet,  Mistral, Math$\Sigma$tral, and Llama 3.1, on tasks designed to measure their propensity for additive versus subtractive modifications. Our findings demonstrate a significant preference for additive changes across all tested models. For example, in a palindrome creation task, Llama 3.1 favored adding letters 97.85\% of the time over removing them. 
Similarly, in a Lego tower balancing task, GPT-3.5 Turbo chose to add a brick 76.38\% of the time rather than remove one. 
In a text summarization task, Mistral 7B  produced longer summaries in 59.40\% to 75.10\% of cases when asked to improve its own or others' writing. These results indicate that, similar to humans, LLMs exhibit a marked additive bias, which might have implications when LLMs are used on a large scale. Addittive bias might increase resource use and environmental impact, leading to higher economic costs due to overconsumption and waste. This bias should be considered in the development and application of LLMs to ensure balanced and efficient problem-solving approaches.

\end{abstract}
\section{Introduction}

Large Language Models (LLMs) present substantial opportunities as tools to aid a growing variety of decision-making processes. However, because they are trained on data generated by humans, LLMs are known to inherit societal biases and can exhibit biases that closely resemble cognitive biases, defined as systematic and erroneous response patterns in judgment and decision-making \citep{tversky1974judgment}. Such human-like biases have the potential to hinder the fairness and transparency of decisions made with the help of LLMs.
Adams et al. \citep{adams2021people} conducted a series of experiments to explore a cognitive phenomenon known as the \textit{addition bias} in human participants. This bias was examined in scenarios where problems could be resolved by either adding or removing elements. Additive transformations result in a state with more elements than the original, while subtractive transformations lead to a state with fewer elements \citep{vergnaud2020classification}. A key finding from Adams et al.'s work was that people tend to add rather than remove elements when modifying ideas, objects, or situations. This tendency was observed across various tasks, such as stabilizing a Lego structure, improving a miniature golf course, creating symmetry within a grid, or rewriting an article summary. Interestingly, participants often chose to add elements even when a subtractive solution would have been simpler and required fewer steps. Additionally, instructions to “improve” a design amplified the addition bias more than instructions to “worsen” a design.
In this paper, we extend this line of inquiry to LLMs, investigating whether they exhibit a similar additive bias. We conducted a series of experiments designed to test the tendency of LLMs to favor additive over subtractive changes across various tasks. These experiments included creating palindromes from strings, balancing Lego towers, modifying recipes with unusual ingredients, improving soup recipes with varying numbers of ingredients, and revising text summaries. We tested several prominent LLMs, including GPT-3.5 Turbo, Claude 3.5 Sonnet, and Llama 3.1 70B, among others. Our research aims to uncover whether these AI models, trained on human-generated data, have inherited the human tendency towards additive problem-solving, and to explore the implications of such a bias for AI-assisted decision-making and problem-solving processes. \\

\textbf{RQ:} \textit{Do Large Language Models exhibit an additive bias similar to humans when solving problems or generating content, and if so, how does this bias manifest across different tasks and models?}

\section{Related works}

LLMs have demonstrated remarkable abilities in various tasks, such as document summarization \citep{wang2023zero}, solving math problems \citep{imani2023mathprompter}, and providing chat support \citep{lee2023prompted}. This has led to their growing use for assistance and advice in daily decision-making \citep{rastogi2023supporting, li2022pre}. However, these models are not immune to algorithmic biases \citep{acerbi2023large}, highlighting the need for strategies to evaluate and mitigate these issues \citep{zhao2018gender, nadeem2020stereoset, liang2021towards}.
During training, LLMs can encode societal biases related to race, gender, and other sensitive areas, potentially generating outputs that reinforce harmful stereotypes or discriminatory views. A common example is the tendency of LLMs to associate certain professions or traits with specific genders or ethnicities, such as linking engineering with male pronouns and nursing with female pronouns \citep{kotek2023gender}. Additionally, biases can surface in text generated on sensitive topics like politics, religion, or social issues \citep{liu2022quantifying, gover2023political, abid2021persistent, liang2021towards}.
In addition to societal bias, LLMs can show answer patterns similar to human-like \textit{cognitive bias} \citep{binz2023using}, which
can implicitly mislead a user’s decision-making \citep{schramowski2022large}.
Cognitive bias refers to a systematic pattern of deviation from norms of rationality in judgment, where individuals create their own “subjective reality” from their perception of the input \citep{haselton2015evolution}, \citep{tversky1982judgment}. 

Three of the eight experiments conducted by \citep{adams2021people} were replicated in \citep{fillon2022people}, confirming the presence of the addition bias. Further, \citep{winter2023more} demonstrated that the addition bias extends beyond behavioral manifestations and is also evident in language. A frequency analysis of the Corpus of Contemporary American English \citep{davies2010corpus} revealed that words associated with increasing quantities, such as \textit{“add”} and \textit{“more”}, are more common in English than those associated with decreasing quantities, such as \textit{“subtract”} and \textit{“less”}.

\section{Methodology}
For the various experiments, the following models were used:
\begin{itemize}
    \setcounter{footnote}{0} % Resetta il contatore delle footnote
    \setlength{\itemsep}{5pt} 
    \item[-] \textbf{GPT-3.5 Turbo}\footnote{\url{https://platform.openai.com/docs/models/gpt-3-5-turbo}}, prompted using the OpenAI API.
    \item[-] \textbf{Claude 3.5 Sonnet}\footnote{\url{https://www.anthropic.com/news/claude-3-5-sonnet}}, prompted using Anthropic API\footnote{\url{https://docs.anthropic.com/en/api/getting-started\#accessing-the-api}}.
    \item[-] \textbf{Math$\Sigma$tral}\footnote{\url{https://mistral.ai/news/mathstral/}}, a model specializing in mathematical and scientific tasks, whose weights were downloaded from HuggingFace\footnote{\url{https://huggingface.co/mistralai/Mathstral-7B-v0.1}}.
    \item[-] \textbf{Llama 3.1 70B} and \textbf{450B}\footnote{\url{https://llama.meta.com/}}, prompted using NVIDIA AI Foundry API\footnote{\url{https://build.nvidia.com/explore/discover}}.
    \item[-] \textbf{Mistral 7B}\footnote{\url{https://mistral.ai/news/announcing-mistral-7b}}, prompted using Mistral AI API\footnote{\url{https://mistral.ai/}}.

\end{itemize}

\begin{figure}
    \centering
    \includegraphics[width=0.6\textwidth]{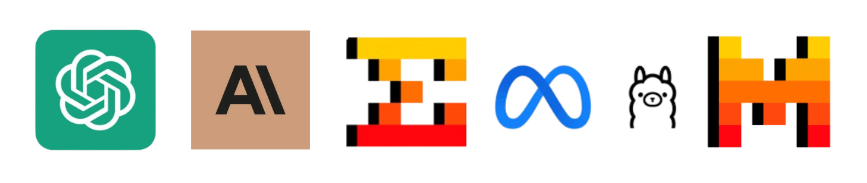} % Include the image with a width of half the text width % Add a caption
    \label{fig:loghi} % Add a label for referencing
\end{figure}

\vspace{12pt}
A \textit{temperature = 0.7} has been set to enhance the variability of the responses, and the following \textit{system prompt} has been used:   
\begin{mycolorbox}
\textit{"Imagine being a regular person asked a question for data collection in a scientific study."}
\end{mycolorbox}

All responses provided by the models were saved in CSV files (available \href{https://github.com/LucaSantagata/More-is-More-Addition-Bias-in-Large-Language-Models}{here}) for result analysis. Initially, as described in greater detail in the following sections, all responses deemed incorrect were excluded, including those that were formally incorrect from a logical standpoint and those that did not meet the prompt's requirements.
For each iteration of the process, the history of previous interactions was neither retained nor used to inform the subsequent generation by the model. Each request was thus treated as an independent instance, without any context from prior interactions.

\section{Experiments}

\subsection{Palindrome sequence task}
In this experiment, the objective was to transform a sequence of letters into a palindrome. Specifically, the strings \textit{"abb"} and \textit{"abab"} were used, along with the following prompts:

\begin{yellowcolorbox}
\textit{"Knowing that a sequence is said to be a palindrome if it is equal to its reverse, or in other words, if reading the sequence from left to right gives the same result as reading it from right to left, you need to make this sequence \textbf{'abb'} a palindrome, but you can only \underline{add or remove} one letter. Give me your answer in one sentence."}
\end{yellowcolorbox}

\begin{yellowcolorbox}
\textit{"Knowing that a sequence is said to be a palindrome if it is equal to its reverse, or in other words, if reading the sequence from left to right gives the same result as reading it from right to left, you need to make this sequence \textbf{'abab'} a palindrome, but you can only \underline{add or remove} one letter. Give me your answer in one sentence."}
\end{yellowcolorbox}

In particular, to prevent the indication \textit{\underline{add or remove}} from influencing the choice by favoring the additive approach over the subtractive one, the experiment was repeated the same number of times for each model using the same prompt, but this time with the inverted instruction \textit{\underline{remove or add}}. All results presented are averages of the values obtained in both cases.

\subsubsection{\textit{GPT-3.5 Turbo} results}

In the case of the sequence \textbf{\textit{"abb"}}, as an initial analysis, we considered only the logically correct responses, where a single letter was used to make it symmetric. These correspond to the answers \textit{"abba"} (where an \textit{"a"} was added at the end) or \textit{"bb"} (where the first \textit{"a"} was removed). Out of the 1000 responses obtained with the suggestion \underline{\textit{add or remove}}, 700 responses were deemed valid, while with \underline{\textit{remove or add}}, 707 responses were considered correct. The results, obtained from the average of the two cases, are presented in Table \ref{tab: abb_GPT}.

\begin{table}[h]
\centering
\renewcommand{\arraystretch}{1.5}
\begin{tabular}{|c|c|c|}
\hline
\multicolumn{2}{|c|}{\textbf{Answer's type (\%)}} \\
\hline
\hline
\textbf{\textit{\quad'abba'\quad}} & \textbf{\textit{'bb'}} \\
\hline
\hline
\textbf{99.50} & \(0.50\)  \\
\hline
\end{tabular}
\caption{\textit{GPT-3.5 Turbo} correct palindrome sequences from \textit{"abb"}}
\label{tab: abb_GPT}
\end{table}

Subsequently, we decided also to consider responses that are not logically correct or do not adhere to using only one letter but are still palindromes (e.g., \textit{"adding the letter 'a' to the middle of the sequence 'abb' to make it a palindrome, resulting in 'abba'"}, where the answer \textit{'abba'} is a palindrome but does not correspond to the given explanation, therefore logically incorrect).
With these new considerations, the number of responses considered correct with the reccomendation \underline{\textit{add or remove}} was 928, while with \underline{\textit{remove or add}} it was 931. Additionally, beyond the two sequences \textit{'abba'} and \textit{'bb'}, which were considered the only correct ones in the previous case, new palindrome sequences emerged, as shown in the results Table \ref{tab: abb_GPT_total}.

\begin{table}[h]
\centering
\renewcommand{\arraystretch}{1.5} 
\begin{tabular}{|c|c|c|c|c|c|c|c|c|}
\hline
\multicolumn{9}{|c|}{\textbf{Answer's type (\%)}} \\  % Riga che copre tutte le colonne
\hline
\hline
\textbf{\textit{'abba'}} & \textbf{\textit{'aba'}} & \textbf{\textit{'bbabb'}} & \textbf{\textit{'bab'}} & \textbf{\textit{'baab'}} & \textbf{\textit{'bb'}} & \textbf{\textit{'abbbba'}} & \textbf{\textit{'abbba'}} & \textbf{\textit{'bbb'}} \\
\hline
\hline
\textbf{93.75} & 2.62 & 1.77 & 0.70 & 0.53 & 0.27 & 0.26 & 0.05 & 0.05 \\
\hline
\end{tabular}
\caption{\textit{GPT-3.5 Turbo} palindrome sequences from \textit{"abb"}}
\label{tab: abb_GPT_total}
\end{table}

On the other hand, the sequence \textbf{”abab”} was tested 100 times for each of the two prompts: with the suggestion \underline{\textit{add or remove}}, 22 responses were discarded as they were neither correct from the palindromic perspective nor the logical one, while with \underline{\textit{remove or add}} 23 responses were discarded for the same reason. Unlike the previous case, all the remaining responses fell into one of the 4 valid solutions that could be obtained by adding or removing only one letter, whose results are shown in Table \ref{tab:abab_GPT}.

\begin{table}[h]
\centering
\renewcommand{\arraystretch}{1.5} 
\begin{tabular}{|c|c|c|c|}
\hline
\multicolumn{4}{|c|}{\textbf{Answer's type (\%)}} \\
\hline
\hline
\textbf{\textit{'ababa'}}& \textbf{\textit{'babab'}} & \textbf{\textit{'bab'}} & \textbf{\textit{'aba'}} \\
\hline
\hline
\textbf{94.87} & 4.48 & 0.65 & 0.00 \\
\hline
\end{tabular}
\caption{\textit{GPT-3.5 Turbo} correct palindrome sequences from \textit{"abab"}}
\label{tab:abab_GPT}
\end{table}

\newpage
\subsubsection{\textit{Claude 3.5 Sonnet} results}

With this model, not only were all responses correct both from a palindromic and a logical perspective, but an extreme tendency towards addition was also observed for both the \textit{\textbf{"abb"}} and \textit{\textbf{"abab"}} sequences, as shown by the results in Tables \ref{tab: abb_Claude} and \ref{tab:abab_Claude}.

\begin{table}[h]
\centering
\renewcommand{\arraystretch}{1.5}
\begin{tabular}{|c|c|c|}
\hline
\multicolumn{2}{|c|}{\textbf{Answer's type (\%)}} \\
\hline
\hline
\textbf{\textit{\quad'abba'\quad}} & \textbf{\textit{'bb'}} \\
\hline
\hline
\textbf{100.00} & \(0.00\)  \\
\hline
\end{tabular}
\caption{\textit{Claude 3.5 Sonnet} palindrome sequences from \textit{"abb"}}
\label{tab: abb_Claude}
\end{table}

\begin{table}[h]
\centering
\renewcommand{\arraystretch}{1.5}
\begin{tabular}{|c|c|c|c|}
\hline
\multicolumn{4}{|c|}{\textbf{Answer's type (\%)}} \\
\hline
\hline
\textbf{\textit{'ababa'}} & \textbf{\textit{'babab'}} & \textbf{\textit{'bab'}}& \textbf{\textit{'aba'}}\\
\hline
\hline
\textbf{100.00} & 0.00 & 0.00 & 0.00 \\
\hline
\end{tabular}
\caption{\textit{Claude 3.5 Sonnet} palindrome sequences from \textit{"abab"}}
\label{tab:abab_Claude}
\end{table}

\subsubsection{\textit{Llama 3.1 405B} results}

For the sequence \textbf{\textit{"abb"}}, out of the 200 responses collected, 19 were discarded because they produced a non-palindromic string. In the results shown in the Table \ref{tab: abb_Llama_405B}, some responses like \textit{"remove the last 'b' to get 'ab', then add an 'a' at the end to get the palindrome 'aba'."} were included, even though they were infrequent and, while accurate, did not adhere to the rule of adding or removing just one letter.

\begin{table}[h]
\centering
\renewcommand{\arraystretch}{1.5}
\begin{tabular}{|c|c|c|}
\hline
\multicolumn{3}{|c|}{\textbf{Answer's type (\%)}} \\
\hline
\hline
\textbf{\textit{'abba'}} & \textbf{\textit{'aba'}} & \textbf{\textit{'bb'}} \\
\hline
\hline
\textbf{97.85} & \(1.56\) & \(0.59\) \\
\hline
\end{tabular}
\caption{\textit{Llama 3.1 405B} palindrome sequences from \textit{"abb"}}
\label{tab: abb_Llama_405B}
\end{table}

\newpage Regarding the sequence \textit{\textbf{"abab"}}, out of the 200 responses, 13 were discarded from the results with the prompt containing \underline{\textit{add or remove}}, and 11 were discarded from those with  \underline{\textit{remove or add}}. In this case as well, all remaining responses fell into one of the 4 valid solutions that could be obtained by adding or removing only one letter, with the results shown in Table \ref{tab: abab_Llama_405B}.

\begin{table}[h]
\centering
\renewcommand{\arraystretch}{1.5}
\begin{tabular}{|c|c|c|c|}
\hline
\multicolumn{4}{|c|}{\textbf{Answer's type (\%)}} \\
\hline
\hline
\textbf{\textit{'ababa'}} & \textbf{\textit{'babab'}} & \textbf{\textit{'aba'}} & \textbf{\textit{'bab'}} \\
\hline
\hline
\textbf{64.17} & \(3.16\) & \(15.01\) & \(17.66\) \\
\hline
\end{tabular}
\caption{\textit{Llama 3.1 405B} palindrome sequences from \textit{"abab"}}
\label{tab: abab_Llama_405B}
\end{table}

\begin{figure}[h!]
    \centering
    \begin{subfigure}[b]{0.45\textwidth}
        \centering
        \includegraphics[width=\linewidth]{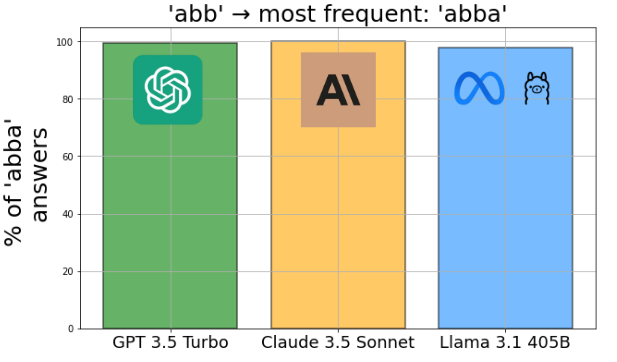}
        \caption{Percentage of times \textit{'abb'} was transformed into \textit{'abba'}}
        \label{fig:a}
    \end{subfigure}\hfill
    \begin{subfigure}[b]{0.45\textwidth}
        \centering
        \includegraphics[width=\linewidth]{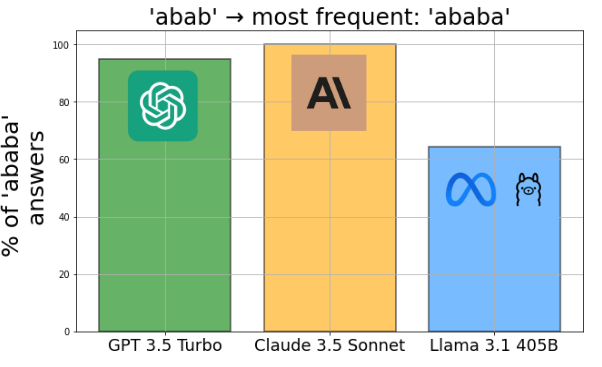}
        \caption{Percentage of times \textit{'abab'} was transformed into \textit{'ababa'}}
        \label{fig:b}
    \end{subfigure}
    \vspace{1em} % Aggiungi spazio verticale qui
    \caption{The two most frequent transformations of \textit{'abb'} and \textit{'abab'} were identified, and their occurrence percentages were plotted}
    \label{fig:Tommaso}
\end{figure}

The presented results clearly show that each model extensively pursued an additive approach. This confirms that the presence of an additive bias has influenced the models' choices, with all of them preferring addition over subtraction.
Moreover, from Figure \ref{fig:Tommaso}, it is also noteworthy that additive responses are not only more frequent but, in both cases and across all models, they occur by a significant margin.\\

\newpage
\subsection{\textit{Lego} towers task}
\begin{figure}
    \centering
    \includegraphics[width=0.5\textwidth]{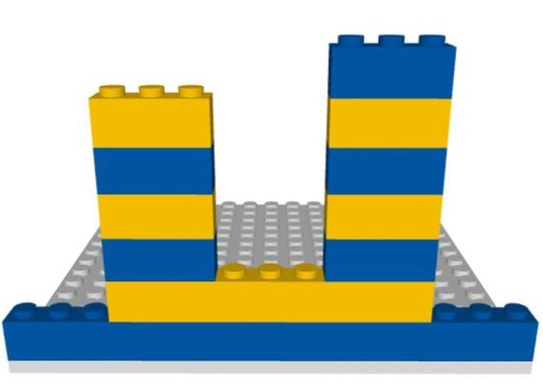} % Include the image with a width of half the text width
    \caption{Example of \textit{Lego} towers construction mentioned in the prompt.} % Add a caption
    \label{fig:lego} % Add a label for referencing
\end{figure}

For this experiment (considere figure \ref{fig:lego} as an example),  it was asked:

\begin{yellowcolorbox}
\textit{"Imagine you have two towers built with Lego bricks. One of them was built with 5 bricks, while the other with 4 bricks. You need to make them the same height using the fewest number of pieces. What do you do? Give me your answer in one sentence."}
\end{yellowcolorbox}

\vspace{12pt}
In this case, using the fewest possible pieces, there are two valid answers to make the towers symmetrical: add one \textit{Lego} brick to the shorter tower on the left \textit{(additive responses)}, or remove one brick from the taller tower on the right \textit{(subtractive responses)}. The purpose of the experiment is to determine if there is an additive
bias in the responses that could explain a majority of answers involving adding a brick rather than
removing one. 
\subsubsection{\textit{GPT-3.5 Turbo} results}
Out of the 1000 responses, 365 were discarded because the result was either logically incorrect or did not result in two symmetrical towers (e.g., \textit{"I would take two bricks from the tower with 5 bricks and add them to the tower with 4 bricks to make both towers have 6 bricks each."}). The results of the remaining 635 correct responses, presented in Table \ref{tab:Lego_ChatGPT}, show that the additive one was the more frequent of the two suggestions.

\begin{table}[h!]
\centering
\renewcommand{\arraystretch}{1.5} 
\begin{tabular}{|c|c|c|}
\hline
\textbf{Additive responses (\%)} & \textbf{Subtractive responses (\%)} \\
\hline
\hline
\textbf{76.38} &  \(23.62\)  \\
\hline
\end{tabular}
\caption{\textit{GPT-3.5 Turbo} choices for the \textit{Lego} experiment}
\label{tab:Lego_ChatGPT}
\end{table}

\subsubsection{\textit{Claude 3.5 Sonnet} results}
For this model, not only were all 1000 collected responses logically correct, but all of them were additive, as indicated by the results in Table \ref{tab:Lego_Claude}.

\begin{table}[h!]
\centering
\renewcommand{\arraystretch}{1.5} 
\begin{tabular}{|c|c|c|}
\hline
\textbf{Additive responses (\%)} & \textbf{Subtractive responses (\%)} \\
\hline
\hline
\textbf{100.00} &  \(0.00\)  \\
\hline
\end{tabular}
\caption{\textit{Claude 3.5 Sonnet} choices for the \textit{Lego} experiment}
\label{tab:Lego_Claude}
\end{table}

\subsubsection{\textit{Math$\Sigma$tral} results}
As in the previous case, for \textit{Math$\Sigma$tral} as well, all 1000 logically correct responses suggested adding a brick to the shorter tower, as indicated in Table \ref{tab:Lego_Mathstral}.

\begin{table}[h!]
\centering
\renewcommand{\arraystretch}{1.5} 
\begin{tabular}{|c|c|c|}
\hline
\textbf{Additive responses (\%)}& \textbf{Subtractive responses (\%)} \\
\hline
\hline
\textbf{100.00} &  \(0.00\)  \\
\hline
\end{tabular}
\caption{\textit{Math$\Sigma$tral} choices for the \textit{Lego} experiment}
\label{tab:Lego_Mathstral}
\end{table}

\subsubsection{\textit{Llama 3.1 70B} results}
Out of the 1000 responses, only 2 were discarded, as they suggested removing the excess brick from the taller tower and placing it on the shorter one, which did not solve the symmetry problem (e.g., \textit{"take one brick from the 5-brick tower and attach it to the 4-brick tower."}).
The results for the remaining 998 answers, shown in Table \ref{tab:Lego_Llama}, unlike the previous model cases, indicate that subtracting a brick from the taller tower was the most frequently proposed solution.

\begin{table}[h!]
\centering
\renewcommand{\arraystretch}{1.5} 
\begin{tabular}{|c|c|c|}
\hline
\textbf{Additive responses (\%)} & \textbf{Subtractive responses (\%)} \\
\hline
\hline
\(1.90\) &  \textbf{98.10}  \\
\hline
\end{tabular}
\caption{\textit{Llama 3.1 70B} choices for the \textit{Lego} experiment}
\label{tab:Lego_Llama}
\end{table}

\vspace{12pt}
A test was conducted to determine whether 485 out of 635 (76.38\%) responses for \textit{GPT-3.5 Turbo}, 1000 out of 1000 (100.00\%) for \textit{Claude 3.5 Sonnet}, 1000 out of 1000 (100\%) for \textit{Math$\Sigma$tral}, and 19 out of 1000 (1.90\%) for \textit{Llama 3.1 70B}, reject the null hypothesis that the suggestions for the two possible transformations are equally likely. The p-value from a two-sided binomial distribution test for these results was found to be less than 0.001.

\begin{figure}
    \centering
    \includegraphics[width=0.8\textwidth]{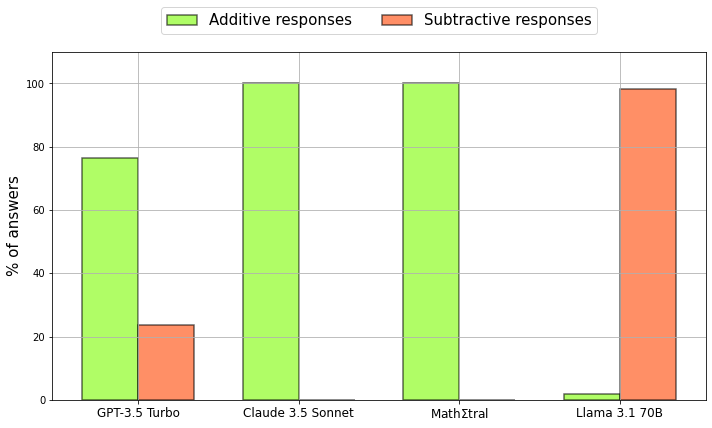} % Include the image with a width of half the text width
    \caption{Models choices for the \textit{Lego} experiment} % Add a caption
    \label{fig:hist_lego} % Add a label for referencing
\end{figure}

\vspace{12pt}
In conclusion, as can be seen also in Figure \ref{fig:hist_lego}, it is possible to confirm that for this type of task, \textit{GPT-3.5 Turbo}, \textit{Claude 3.5 Sonnet}, and \textit{Math$\Sigma$tral} showed a strong additive bias, unlike \textit{Llama 3.1 70B}, which in this case demonstrated a pronounced tendency towards subtractive choices. \\

\subsection{Elementary operation task}
In this experiment, the following prompt was used: 
\begin{yellowcolorbox}
\textit{"Given these numbers: \(\boldsymbol{[n_1, n_2]}\), which of the four basic operations would you suggest performing? Provide your answer in one word."}
\end{yellowcolorbox}

Where \(\boldsymbol{[n_1, n_2]}\) are two numbers within the range of 1 to 10, randomly generated in each of the 1000 iterations. The aim is to investigate the potential tendency to prefer elementary operations that increase the value of the numbers involved, such as addition and multiplication (referred to as \textit{additive operations}), over those that decrease it, such as subtraction and division (referred to as \textit{subtractive operations}).

\subsubsection{\textit{GPT-3.5 Turbo} results}
Out of the 1000 collected responses,
2 were discarded as they suggested \textit{average} rather than one of the four basic operations.
For the remaining 998 responses, the distribution of choices is showed in Table \ref{tab:math_GPT}. 

\begin{table}[h!]
\centering
\renewcommand{\arraystretch}{1.5} 
\begin{tabular}{|c|c|c|c|}
\hline
\textbf{Addition (\%)} & \textbf{Multiplication (\%)} & \textbf{Subtraction (\%)} & \textbf{Division (\%)} \\
\hline
\hline
\textbf{59.02} & 21.74 & 16.43 & 2.81 \\
\hline
\end{tabular}
\caption{\textit{GPT-3.5 Turbo} choices of the elementary operations}
\label{tab:math_GPT}
\end{table}

\subsubsection{\textit{Claude 3.5 Sonnet} results}
In this case, all the responses suggested one of the four basic operations, which is why none of them were discarded. The obtained results are shown in Table \ref{tab:math_Claude}.

\begin{table}[h!]
\centering
\renewcommand{\arraystretch}{1.5} 
\begin{tabular}{|c|c|c|c|}
\hline
\textbf{Addition (\%)} & \textbf{Multiplication (\%)} & \textbf{Subtraction (\%)} & \textbf{Division (\%)} \\
\hline
\hline
\textbf{46.20} & 19.80 & 33.30& 0.70 \\
\hline
\end{tabular}
\caption{\textit{Claude 3.5 Sonnet} choices of the elementary operations}
\label{tab:math_Claude}
\end{table}

\subsubsection{\textit{Math$\Sigma$tral} results}

Also in this case it was not necessary to remove any responses, and the results obtained, shown in Table \ref{tab:math_Mathstral}, indicate that the only operations suggested were addition and multiplication, with no indication of subtraction or division. 

\begin{table}[h!]
\centering
\renewcommand{\arraystretch}{1.5} 
\begin{tabular}{|c|c|c|c|}
\hline
\textbf{Addition (\%)} & \textbf{Multiplication (\%)} & \textbf{Subtraction (\%)} & \textbf{Division (\%)} \\
\hline
\hline
15.30 & \textbf{84.70} & 0.00 & 0.00 \\
\hline
\end{tabular}
\caption{\textit{Math$\Sigma$tral} choices of the elementary operations}
\label{tab:math_Mathstral}
\end{table}

\subsubsection{\textit{Llama 3.1 70B} results}
All 1000 responses suggested one of the four basic operations, which is why they were all included in the analysis presented in Table \ref{tab:math_Llama}. The results demonstrate a clear tendency for the model to favor \textit{addition}.

\begin{table}[h!]
\centering
\renewcommand{\arraystretch}{1.5} 
\begin{tabular}{|c|c|c|c|}
\hline
\textbf{Addition (\%)} & \textbf{Multiplication (\%)} & \textbf{Subtraction (\%)} & \textbf{Division (\%)} \\
\hline
\hline
\textbf{63.50} & 5.10 & 20.30 & 10.90 \\
\hline
\end{tabular}
\caption{\textit{Llama 3.1 70B} choices of the elementary operations}
\label{tab:math_Llama}
\end{table}

A test was conducted to determine whether 753 (sum of addition and multiplication counts) out of 998 responses (75.45\%) for \textit{GPT-3.5 Turbo}, 660 out of 1000 (66.00\%) for \textit{Claude 3.5 Sonnet}, 1000 out of 1000 (100.00\%) for \textit{Math$\Sigma$tral}, and 686 out of 1000 (68.60\%) for \textit{Llama 3.1 70B}, reject the null hypothesis that suggestions for \textit{additive operations} are equally likely as those for \textit{subtractive operations}. The p-value from a two-sided binomial distribution test for these results was found to be less than 0.001. 
\vspace{12pt}

\begin{figure}
    \centering
    \includegraphics[width=1\textwidth]{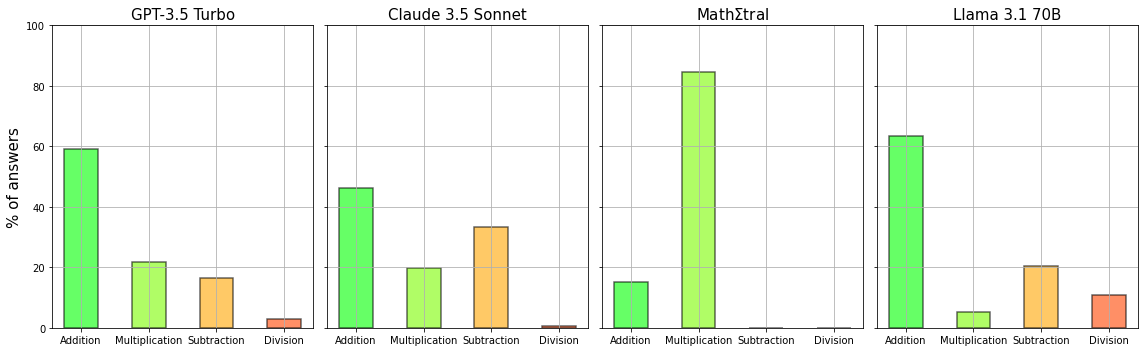} % Include the image with a width of half the text width
    \caption{Models choices of the elementary operations} % Add a caption
    \label{fig:riassunto_math} % Add a label for referencing
\end{figure}

As summarized in Figure \ref{fig:riassunto_math}, all models suggested addition and multiplication more frequently than subtraction and division, clearly indicating the presence of an additive bias for this type of task. \\

\newpage
\subsection{Anomalous sandwich task}
The objective of this experiment is to determine if the tested LLMs are more likely to add or subtract from stimuli with anomalous components. To test this hypothesis, it was asked to modify the recipe for a cheese sandwich. Specifically, the prompt used was as follows:
\begin{yellowcolorbox}
\textit{"Imagine you are hungry and decide to make a sandwich for lunch. Below there is a list of five ingredients: bread, ham, cheese, lettuce, mayonnaise, and \textbf{ingredient n°6}. In one sentence, please describe how you would change this recipe when making your sandwich."}
\end{yellowcolorbox}

For \textit{\textbf{ingredient n°6}}, three different possibilities were tested: \\
\textit{banana}, \textit{chocolate}, and \textit{pineapple}. These three ingredients are extremely unusual for a cheese sandwich recipe, which is why it would be reasonable to expect that the most obvious response would be of a reductive nature, namely only the removal of the sixth unusual ingredient. However, the results demonstrated that this is not so straightforward.

The collected responses were divided into four categories:
\begin{itemize}
    \item[-] \textbf{no change}: the original sandwich recipe is left unaltered with the unusual ingredient (e.g \textit{"I would make a ham and cheese sandwich with lettuce, mayonnaise, and banana slices for a unique twist"}).
    \item[-] \textbf{only addition}: there is a modification of only additive type, that is, the addition of an ingredient, but no subtractive modification, that is, the removal of an ingredient (e.g \textit{"If it were up to me, I would add some sliced tomatoes to my sandwich for an extra burst of freshness and flavor"}).
    \item[-] \textbf{only remotion}: there is a modification of only subtractive type, but no additive modification (e.g \textit{"I would remove the chocolate from the list of ingredients when making my sandwich"}).
    \item[-] \textbf{both addition and remotion}: there are simultaneously both an additive modification and a subtractive modification, indicating both the removal and the addition of an ingredient (e.g. \textit{"I would skip the pineapple and add some mustard for an extra kick of flavor in my sandwich"}).
\end{itemize}

\subsubsection{\textit{GPT-3.5 Turbo results} results}

As mentioned earlier, when faced with an extremely unusual ingredient, the most reasonable choice would be to remove it. However, the results in Table \ref{tab:sandwich_GPT} show a marked additive tendency of the model. Instead of simply removing the unusual ingredient, \textit{GPT-3.5 Turbo} frequently suggested adding a new ingredient. This was the most common choice in the cases of \textit{banana} and \textit{pineapple}, and while it was not the most frequent choice in the case of \textit{chocolate}, it was still suggested a significantly notable number of times.

\begin{table}[h]
\centering
\renewcommand{\arraystretch}{1.5} 
\begin{tabular}{|c|c|c|c|}
\hline
\textbf{Answer's type} & \textbf{Banana (\%)} & \textbf{Chocolate (\%)} & \textbf{Pineapple (\%)} \\
\hline
\hline
no change&  \(0.20\) & \(0.00\) & \(0.00\) \\
\hline
\hline
only addition &  \(0.20\) &  \(0.00\) &  \(1.00\) \\
\hline
\hline
only remotion &  \(36.30\) & \(\textbf{54.70}\) &  \(25.38\) \\
\hline
\hline
both addition and remotion & \(\textbf{63.30}\) &  \(45.30\)  & \(\textbf{73.62}\) \\
\hline
\end{tabular}
\caption{\textit{GPT-3.5} choices for each anomalous ingredient}
\label{tab:sandwich_GPT}
\end{table}

\subsubsection{\textit{Claude 3.5 Sonnet} results}

The responses from \textit{Claude 3.5 Sonnet}, shown in Table \ref{tab:sandwich_Claude}, generally met the expectation of simply removing the unusual ingredient, especially in the cases of \textit{banana} and \textit{chocolate}. However, it is interesting to note that in the case of \textit{pineapple}, although the suggestion to remove only is the most frequent, a significant percentage of responses included the addition of a new ingredient.

\begin{table}[h]
\centering
\renewcommand{\arraystretch}{1.5} 
\begin{tabular}{|c|c|c|c|}
\hline
\textbf{Answer's type} & \textbf{Banana (\%)} & \textbf{Chocolate (\%)} & \textbf{Pineapple (\%)} \\
\hline
\hline
no change & \(0.00\) &  \(0.00\) &  \(0.00\) \\
\hline
\hline
only addition &  \(0.00\) &  \(0.00\) &  \(0.00\) \\
\hline
\hline
only remotion & \textbf{99.90} & \textbf{96.80} & \textbf{71.70} \\
\hline
\hline
both addition and remotion &  \(0.10\) &  \(3.20\) &  \(28.30\) \\
\hline
\end{tabular}
\caption{\textit{Claude 3.5 Sonnet} choices for each anomalous ingredient}
\label{tab:sandwich_Claude}
\end{table}

\subsubsection{\textit{Llama 3.1 70B} results}

Table \ref{tab:sandwich_Llama} shows that \textit{Llama 3.1 70B}, for each individual ingredient and in all cases, consistently chose the most straightforward solution, which was to simply remove the unusual component.

\begin{table}[h]
\centering
\renewcommand{\arraystretch}{1.5} 
\begin{tabular}{|c|c|c|c|}
\hline
\textbf{Answer's type} & \textbf{Banana (\%)} & \textbf{Chocolate (\%)} & \textbf{Pineapple (\%)} \\
\hline
\hline
no change & \(0.00\) & \(0.00\) & \(0.00\) \\
\hline
\hline
only addition & \(0.00\) & \(0.00\) & \(0.00\) \\
\hline
\hline
only remotion & \textbf{100.00} & \textbf{100.00} & \textbf{100.00} \\
\hline
\hline
both addition and remotion & \(0.00\) & \(0.00\) & \(0.00\) \\
\hline
\end{tabular}
\caption{\textit{Llama 3.1 70B} choices for each anomalous ingredient}
\label{tab:sandwich_Llama}
\end{table}

\begin{figure}
    \centering
    \includegraphics[width=1\textwidth]{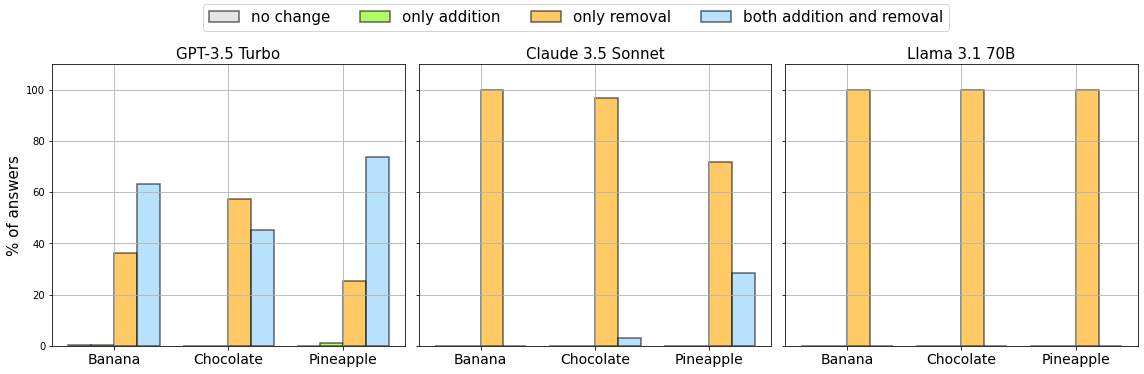} % Include the image with a width of half the text width
    \caption{Models' choices for the anomalous ingredient} % Add a caption
    \label{fig:hist_sandwich} % Add a label for referencing
\end{figure}

In conclusion, as shown in Figure \ref{fig:hist_sandwich}, in this type of task, only \textit{GPT-3.5 Turbo} displayed an additive bias. This was confirmed by the interesting fact that it did not merely remove the unwanted ingredient but often preferred to suggest adding a new one as well. \\

\newpage
\subsection{Increasing ingredients in soup task}
For this task, it was asked to transform a soup recipe using the following prompt: 
\begin{yellowcolorbox}
\textit{"Below you have a list of ingredients for a soup recipe: \underline{\{ingredients\}}. Your job is to make any and all changes necessary to improve this soup. Assume that this soup is for someone who has no dietary restrictions or strong food dislikes. Please provide your answer in only one sentence."}
\end{yellowcolorbox}

where the number of \textit{\underline{\{ingredients\}}} was increased each time, covering the following cases:

\begin{itemize}
    \item[-] \textbf{5 ingredients}: vegetable broth, carrots, peas, garlic, salt/pepper.
    \item[-] \textbf{15 ingredients}: vegetable broth, carrots, peas, garlic, salt/pepper, onion, celery, oregano, potatoes, thyme, green beans, corn, zucchini, parsley, and leeks.
    \item[-] \textbf{30 ingredients}: vegetable broth, carrots, peas, garlic, salt/pepper, onion, celery, oregano, potatoes, thyme, green beans, corn, zucchini, parsley, leeks, tomatoes, spinach, bell peppers, mushrooms, lentils, cabbage, chickpeas, bay leaves, paprika, cumin, lemon juice, ginger, cilantro, basil, kale.
    \item[-] \textbf{50 ingredients}: vegetable broth, carrots, peas, garlic, salt, pepper, onion, celery, oregano, potatoes, thyme, green beans, corn, zucchini, parsley, leeks, tomatoes, spinach, bell peppers, mushrooms, lentils, cabbage, chickpeas, bay leaves, paprika, cumin, lemon juice, ginger, cilantro, basil, kale, cauliflower, green onions, black beans, quinoa, broccoli, radishes, fennel, mint, dill, rosemary, sage, tofu, coconut milk, turmeric, chili powder, sweet potatoes, barley, shallots, pumpkin, asparagus, lime juice.
\end{itemize}

The goal of this experiment is to investigate whether there is a tendency for models to add ingredients rather than remove them. Specifically, it aims to observe how this tendency might be influenced by the increasing number of ingredients and whether there is a "phase transition" where this tendency is no longer observed. 

For this reason, the responses were categorized as follows:

\begin{itemize}
\item[-] \textbf{Only addition:} A suggestion was made to add one or more ingredients (e.g., \textit{"I would add some diced potatoes and onion to enhance the flavor and texture of the soup"}).
\item[-] \textbf{Only removal:} A suggestion was made to remove one or more ingredients (e.g., \textit{"I would remove the radishes and fennel, as they may overpower the other flavors in the soup"}).
\item[-] \textbf{Both addition and removal:} A suggestion was made to remove one or more elements while also suggesting the addition of new ones (e.g., \textit{"I would remove the barley and pumpkin and add in a dash of smoked paprika and a splash of balsamic vinegar for a richer flavor profile"}).
\end{itemize}

\subsubsection{\textit{GPT-3.5 Turbo} results}

As shown by the results in Table \ref{tab:Soup_GPT}, with 5 and 15 ingredients, almost every suggestion from \textit{GPT-3.5 Turbo} is additive. It takes 30 ingredients before suggestions that involve removing an ingredient start to appear consistently, although the additive tendency remains the most prevalent overall. With 50 ingredients, a reasonably high number, the most common suggestion shifts to removal. However, surprisingly, the difference is not excessive, and the additive tendency still appears in responses that include both types of suggestions as well as in those that are exclusively additive. 

\begin{table}[h]
\centering
\renewcommand{\arraystretch}{1.5} 
\begin{tabular}{|c|c|c|c|}
\hline
\multicolumn{1}{|c|}{} & \multicolumn{3}{c|}{\textbf{Answer's type (\%)}} \\
\cline{1-4}
\hline
\hline
\textbf{n° of ingredients} & \textbf{Only addition } & \textbf{Only remotion } & \textbf{Both add. and rem.} \\
\hline
\hline
5 & \(\textbf{100.00}\) &  \(0.00\)  &  \(0.00\) \\
\hline
\hline
15 & \(\textbf{99.70}\) &  \(0.00\) &  \(0.30\) \\
\hline
\hline
30 & \(\textbf{76.91}\) &  \(12.15\) & \(10.94\) \\
\hline
\hline
50 &  \(27.50\)& \(\textbf{ 37.60}\) &  \(34.90\) \\
\hline
\end{tabular}
\caption{\textit{GPT-3.5 Turbo} choices by ingredients number}
\label{tab:Soup_GPT}
\end{table}

\subsubsection{\textit{Claude 3.5 Sonnet} results}

In this case, as shown in Table \ref{tab:Soup_Claude}, with 5 and 15 ingredients, the suggestions are exclusively additive, proposing to add one or more ingredients. Even with 30 ingredients, the trend remains largely the same, with suggestions involving only removal being completely absent. With 50 ingredients, this trend reverses, but surprisingly, removal is still not the preferred solution. Instead, the preferred suggestion is a combination of both removal and addition. This confirms that for this model, a persistent additive tendency is consistently present in this type of task.

\begin{table}[h]
\centering
\renewcommand{\arraystretch}{1.5} 
\begin{tabular}{|c|c|c|c|}
\hline
\multicolumn{1}{|c|}{} & \multicolumn{3}{c|}{\textbf{Answer's type (\%)}} \\
\cline{1-4}
\hline
\hline
\textbf{n° of ingredients} & \textbf{Only addition } & \textbf{Only remotion } & \textbf{Both add. and rem.} \\
\hline
\hline
5 & \(\textbf{100.00}\) & \(0.00\) &  \(0.00\) \\
\hline
\hline
15 & \(\textbf{100.00}\) &  \(0.00\) &  \(0.00\) \\
\hline
\hline
30 & \(\textbf{93.20}\) &  \(0.00\) &  \(6.80\) \\
\hline
\hline
50 & \(2.50\) &  \(10.70\) & \(\textbf{86.80}\) \\
\hline
\end{tabular}
\caption{\textit{Claude 3.5 Sonnet} choices by ingredients number}
\label{tab:Soup_Claude}
\end{table}

\subsubsection{\textit{Llama 3.1 70B} results}
As indicated by the results in Table \ref{tab:Soup_Llama}, with 5 ingredients, as usual, the model tends to suggest adding. However, as the number of ingredients increases, the strategy of both removing and adding becomes increasingly frequent, eventually becoming the preferred option with 50 ingredients, in contrast to single removals, which only reach a consistent percentage in the scenario with the highest number of ingredients. Even in this case, it is possible to observe a strong additive bias that favors the addition of elements over their removal.

\begin{table}[h]
\centering
\renewcommand{\arraystretch}{1.5} 
\begin{tabular}{|c|c|c|c|}
\hline
\multicolumn{1}{|c|}{} & \multicolumn{3}{c|}{\textbf{Answer's type (\%)}} \\
\cline{1-4}
\hline
\hline
\textbf{n° of ingredients} & \textbf{Only addition } & \textbf{Only remotion } & \textbf{Both add. and rem.} \\
\hline
5 & \(\textbf{99.90}\) & \(0.00\) &  \(0.10\) \\
\hline
\hline
15 & \(\textbf{66.50}\) &  \(0.60\) &  \(32.90\) \\
\hline
\hline
30 & \(0.00\) &  \(10.70\) &  \(\textbf{89.30}\) \\
\hline
\hline
50 & \(0.00\) &  \(28.90\) & \(\textbf{71.10}\) \\
\hline
\end{tabular}
\caption{\textit{Llama 3.1 70B} choices by ingredients number}
\label{tab:Soup_Llama}
\end{table}

\begin{figure}
    \centering
    \includegraphics[width=1\textwidth]{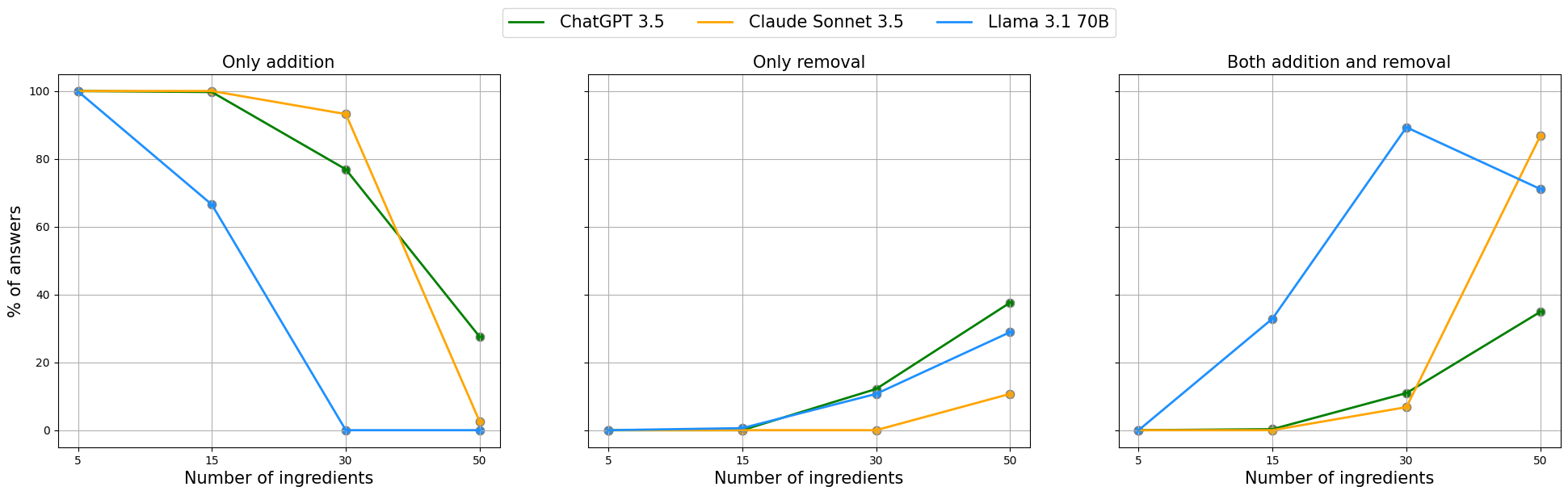} % Include the image with a width of half the text width
    \caption{Models' choices by ingredients number} % Add a caption
    \label{fig:soup_plot} % Add a label for referencing
\end{figure}

\vspace{12 pt}
Examining Figure \ref{fig:soup_plot}, it is evident that the response patterns are quite similar across all three models, particularly in the case of purely additive responses (which decrease as the number of ingredients increases) and those focused solely on removal (which become more frequent). Notably, in some instances—especially when only 5 ingredients are involved—additive responses make up the entirety of the suggestions, a situation never observed with subtractive responses. Even with 50 ingredients, subtractive responses never reach totality unless combined with an additive suggestion. These findings highlight that all models exhibit a strong additive bias when performing this specific task. \\

\newpage
\subsection{Summarization task}
This experiment allowed the observation of potential additive or subtractive tendencies in the context of a revision process.
During the first phase, each model was asked 1000 times to summarize a text provided in the prompt (specifically the proposed text was the introduction of Wikipedia's page on the Roman Empire\footnote{\url{https://en.wikipedia.org/wiki/Roman_Empire}}, attached in Appendix \ref{app:Summary}).

\begin{yellowcolorbox}
\textit{"Summarize the following text: The Roman Empire was... "}
\end{yellowcolorbox}

Of the provided summary, the number of words used was counted, and then, during the second phase, following the same procedure proposed to the participants by \citep{adams2021people}, it was asked to improve it, but in two different contexts.\\

\textbf{First case - one's own writing} \\
In this case, it was asked to improve the previous summary, presented again in the prompt, specifying that it was a summary elaborated by the model itself.

\begin{yellowcolorbox}
\textit{"Edit your summary with the goal of improving how well you summarized the text."}
\end{yellowcolorbox}

Finally, the number of words used was counted.\\

\textbf{Second case - others' writing} \\
This time, the original summary was always presented with the request to improve it, but with the specification that it was done by someone else other than the model.

\begin{yellowcolorbox}
\textit{"Edit this previous summary made by someone else with the goal of improving how well the text has been summarized."}
\end{yellowcolorbox}

In this case as well, the number of words used in the improved summary was counted. \\

The key observation of this experiment is to determine whether the summaries obtained have more or fewer words than the original summary, thereby highlighting the presence or absence of an additive trend.

\subsubsection{\textit{GPT-3.5 Turbo} results}
As shown in the Table \ref{tab:Summary_GPT}, especially in the first case, the model most often preferred to adopt a subtractive strategy, using fewer words.
\begin{table}[h!]
\centering
\renewcommand{\arraystretch}{1.5} 
\begin{tabular}{|c|c|c|c|}
\hline
\multicolumn{4}{|c|}{\textbf{Answer's type (\%)}} \\ 
\hline
\hline
\multicolumn{2}{|c|}{\textbf{One’s Own Writing (\%)}} & \multicolumn{2}{c|}{\textbf{Others’ Writing (\%)}} \\ 
\hline
\hline
More words & Less words & More words & Less words \\  
\hline
\hline
1.60 & \textbf{98.40} & 38.40 & \textbf{61.60} \\ 
\hline
\end{tabular}
\caption{Percentage of times the \textit{GPT-3.5 Turbo} edited summary contained more or fewer words than the original one}
\label{tab:Summary_GPT}
\end{table}

\subsubsection{\textit{Claude 3.5 Sonnet} results}
Also in this case, as indicated in Table \ref{tab:Summary_Claude} the preferred strategy in both instances was to use fewer words by writing shorter summaries.
\begin{table}[h!]
\centering
\renewcommand{\arraystretch}{1.5} 
\begin{tabular}{|c|c|c|c|}
\hline
\multicolumn{4}{|c|}{\textbf{Answer's type (\%)}} \\ 
\hline
\hline
\multicolumn{2}{|c|}{\textbf{One’s Own Writing (\%)}} & \multicolumn{2}{c|}{\textbf{Others’ Writing (\%)}} \\ 
\hline
\hline
More words & Less words & More words & Less words \\  
\hline
\hline
40.80 & \textbf{59.20} & 39.40 & \textbf{60.60} \\ 
\hline
\end{tabular}
\caption{Percentage of times the \textit{Claude 3.5 Sonnet} edited summary contained more or fewer words than the original one}
\label{tab:Summary_Claude}
\end{table}

\newpage
\subsubsection{\textit{Mistral 7B} results}
Unlike the previous cases, in this instance, the Table \ref{tab:Summary_Mistral} shows that the tendency for both cases was additive, resulting in edited summaries with an higher number of words.
\begin{table}[h!]
\centering
\renewcommand{\arraystretch}{1.5} 
\begin{tabular}{|c|c|c|c|}
\hline
\multicolumn{4}{|c|}{\textbf{Answer's type (\%)}} \\ 
\hline
\hline
\multicolumn{2}{|c|}{\textbf{One’s Own Writing (\%)}} & \multicolumn{2}{c|}{\textbf{Others’ Writing (\%)}} \\ 
\hline
\hline
More words & Less words & More words & Less words \\ 
\hline
\hline
\textbf{59.40} & 40.60 & \textbf{75.10} & 24.90 \\ 
\hline
\end{tabular}
\caption{Percentage of times the \textit{Mistral 7B} edited summary contained more or fewer words than the original one}
\label{tab:Summary_Mistral}
\end{table}

\begin{figure}
    \centering
    \includegraphics[width=1\textwidth]{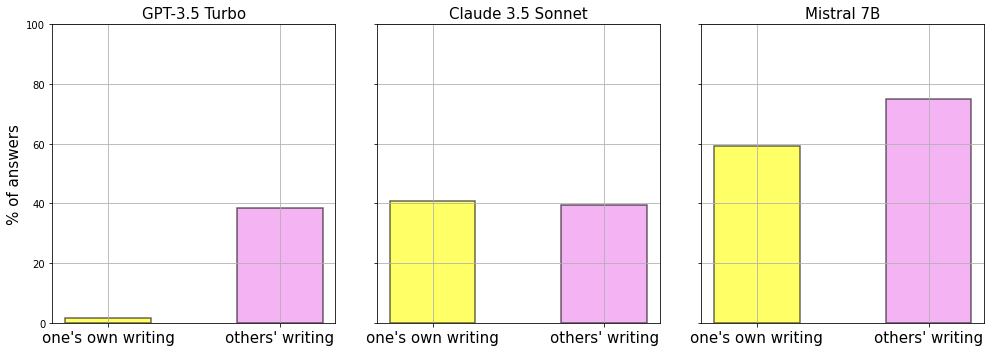} % Include the image with a width of half the text width
    \caption{Percentage of times models' edited summary contained more words than the original one} % Add a caption
    \label{fig:hist_summary} % Add a label for referencing
\end{figure}

\vspace{12 pt}
A test was conducted for both cases to determine if the counts of 16 out of 1000 and 384 out of 1000 for \textit{GPT-3.5 Turbo} (408 out of 1000 and 394 out of 1000 for \textit{Claude 3.5 Sonnet}, and 594 out of 1000 and 751 out of 1000 for \textit{Mistral 7B}) reject the null hypothesis that the probability of producing a summary with fewer words is the same as producing one with more words compared to the original.
The p-value from a two-sided binomial distribution test for these
results was found to be less than 0.001.
As shown in Figure \ref{fig:hist_summary}, it can be concluded that during the revision phase, only \textit{Mistral 7B} exhibited an additive bias in both cases. In contrast, \textit{Claude 3.5 Sonnet} and \textit{GPT-3.5 Turbo} (for the first time in this series of proposed experiments) favored a reductive strategy.  \\

\section{Conclusion}
This study investigated the presence of addition bias in Large Language Models (LLMs), drawing parallels to the cognitive bias observed in humans where individuals tend to favor additive over subtractive changes. Through a series of controlled experiments across various tasks, we tested several prominent LLMs, including GPT-3.5 Turbo, Claude 3.5 Sonnet, Mistral, Math$\Sigma$tral, and Llama 3.1.
Our findings consistently demonstrated a significant preference for additive changes across most tested models and tasks. This bias was particularly evident in tasks such as palindrome creation, Lego tower balancing, and elementary operation selection. For instance, in the palindrome task, models like GPT-3.5 Turbo and Claude 3.5 Sonnet showed a strong tendency to add letters rather than remove them. Similarly, in the Lego tower task, most models preferred adding a brick to the shorter tower rather than removing one from the taller tower.
The addition bias persisted even in more complex scenarios, such as modifying recipes with unusual ingredients or improving soup recipes with varying numbers of ingredients. Interestingly, the bias remained present but decreased in intensity as the number of ingredients increased, suggesting a potential "phase transition" point where subtractive changes become more prevalent.
However, it is important to note that the bias was not uniform across all tasks and models. For example, in the text summarization task, some models (GPT-3.5 Turbo and Claude 3.5 Sonnet) showed a tendency towards reduction rather than addition when improving their own or others' summaries. This suggests that the manifestation of addition bias may be task-dependent and can vary across different LLMs.
These findings have significant implications for the development and application of LLMs in various domains. The presence of addition bias could influence how these models approach problem-solving, decision-making, and content generation tasks. It may lead to unnecessarily complex solutions, inefficiencies, or in certain scenarios it may go against the Occam Principle where simpler, subtractive approaches might be more appropriate. When LLMs are used on a large scale, addition bias can even increase resource use, leading to higher economic costs and increased environmental impact due to overconsumption and waste.
Future research should focus on understanding the root causes of this bias in LLMs, potentially exploring its relationship to training data and model architectures. Additionally, developing strategies to mitigate this bias could be crucial for improving the efficiency and effectiveness of LLMs across a wide range of applications.

\bibliographystyle{abbrvnat}
\bibliography{ref}

\newpage
\begin{appendix}
\renewcommand{\thesection}{\Alph{section}}
\section{Roman Empire text}
\label{app:Summary}
\textit{"The Roman Empire was the post-Republican state of ancient Rome. It is generally understood to mean the period and territory ruled by the Romans following Octavian's assumption of sole rule under the Principate in 27 BC. It included territories in Europe, North Africa, and Western Asia and was ruled by emperors. The fall of the Western Roman Empire in 476 AD conventionally marks the end of classical antiquity and the beginning of the Middle Ages.
Rome had expanded its rule to most of the Mediterranean and beyond. However, it was severely destabilized in civil wars and political conflicts which culminated in the victory of Octavian over Mark Antony and Cleopatra at the Battle of Actium in 31 BC, and the subsequent conquest of the Ptolemaic Kingdom in Egypt. In 27 BC, the Roman Senate granted Octavian overarching power (\textit{imperium}) and the new title of Augustus, marking his accession as the first Roman emperor of a monarchy with Rome as its sole capital. The vast Roman territories were organized in senatorial and imperial provinces.
The first two centuries of the Empire saw a period of unprecedented stability and prosperity known as the \textit{Pax Romana} (lit.~'Roman Peace'). Rome reached its greatest territorial expanse under Trajan (r.~98–117 AD); a period of increasing trouble and decline began under Commodus (r.~180–192). In the 3rd century, the Empire underwent a crisis that threatened its existence, as the Gallic and Palmyrene Empires broke away from the Roman state, and a series of short-lived emperors led the Empire. It was reunified under Aurelian (r.~270–275). Diocletian set up two different imperial courts in the Greek East and Latin West in 286; Christians rose to power in the 4th century after the Edict of Milan. The imperial seat moved from Rome to Byzantium in 330, renamed Constantinople after Constantine the Great. The Migration Period, involving large invasions by Germanic peoples and by the Huns of Attila, led to the decline of the Western Roman Empire. With the fall of Ravenna to the Germanic Herulians and the deposition of Romulus Augustus in 476 AD by Odoacer, the Western Roman Empire finally collapsed. The Eastern Roman Empire survived for another millennium with Constantinople as its sole capital, until the city's fall in 1453.
Due to the Empire's extent and endurance, its institutions and culture had a lasting influence on the development of language, religion, art, architecture, literature, philosophy, law, and forms of government across its territories. Latin evolved into the Romance languages while Medieval Greek became the language of the East. The Empire's adoption of Christianity resulted in the formation of medieval Christendom. Roman and Greek art had a profound impact on the Italian Renaissance. Rome's architectural tradition served as the basis for Romanesque, Renaissance and Neoclassical architecture, influencing Islamic architecture. The rediscovery of classical science and technology (which formed the basis for Islamic science) in medieval Europe contributed to the Scientific Renaissance and Scientific Revolution. Many modern legal systems, such as the Napoleonic Code, descend from Roman law. Rome's republican institutions have influenced the Italian city-state republics of the medieval period, the early United States, and modern democratic republics."}
\end{appendix}

\end{document}